\documentclass[10pt,twocolumn,letterpaper]{article}

\usepackage{iccv}              

\definecolor{iccvblue}{rgb}{0.21,0.49,0.74}
\usepackage[pagebackref,breaklinks,colorlinks,allcolors=iccvblue]{hyperref}

\usepackage{algorithm}
\usepackage{algorithmic}
\usepackage[accsupp]{axessibility}  

\usepackage{pifont}
\newcommand{\cmark}{\textcolor[rgb]{0.0, 0.5, 0.0}{\ding{51}}}
\newcommand{\xmark}{\textcolor{red}{\ding{55}}}

\usepackage{booktabs}
\usepackage{multirow}

\newcommand{\myparagraph}[1]{\vspace{0pt}\noindent{\bf #1}}

\newcommand{\aacc}{$\mathbf{\textit{aAcc}}$\xspace}
\newcommand{\gacc}{$\mathbf{\textit{gAcc}}$\xspace}

\newcommand{\iacc}{the incremental accuracy\xspace}

\newcommand{\refword}[1]{\textcolor[rgb]{0.20, 0.44, 0.69}{#1}}

\def\paperID{11} 
\def\confName{ICCV}
\def\confYear{2025}

\title{
Does Prior Data Matter? \\ Exploring Joint Training in the Context of Few-Shot Class-Incremental Learning
}

\author{
\textbf{Shiwon Kim}$^{1}$\thanks{Equal contribution.} \qquad \textbf{Dongjun Hwang}$^{2*}$\thanks{Corresponding author.} \qquad \textbf{Sungwon Woo}$^{2*}$ \qquad \textbf{Rita Singh}$^{3}$\textsuperscript{\textdagger} \vspace{.5em} \\
{\normalsize
$^1$Yonsei University \xspace\xspace\xspace
$^2$Sogang University \xspace\xspace\xspace
$^3$Carnegie Mellon University
}
\\
{\small\texttt{shiwon1998@yonsei.ac.kr, \{djhwang, swwoo\}@sogang.ac.kr, rsingh@cs.cmu.edu}}
}

\begin{document}

\maketitle

\begin{abstract}
Class-incremental learning (CIL) aims to adapt to continuously emerging new classes while preserving knowledge of previously learned ones. Few-shot class-incremental learning (FSCIL) presents a greater challenge that requires the model to learn new classes from only a limited number of samples per class. While incremental learning typically assumes restricted access to past data, it often remains available in many real-world scenarios. This raises a practical question: should one retrain the model on the full dataset (i.e., joint training), or continue updating it solely with new data? In CIL, joint training is considered an ideal benchmark that provides a reference for evaluating the trade-offs between performance and computational cost. However, in FSCIL, joint training becomes less reliable due to severe imbalance between base and incremental classes. This results in the absence of a practical baseline, making it unclear which strategy is preferable for practitioners. To this end, we revisit joint training in the context of FSCIL by incorporating imbalance mitigation techniques, and suggest a new imbalance-aware joint training benchmark for FSCIL. We then conduct extensive comparisons between this benchmark and FSCIL methods to analyze which approach is most suitable when prior data is accessible. Our analysis offers realistic insights and guidance for selecting training strategies in real-world FSCIL scenarios. Code is available at: {\small\url{https://github.com/shiwonkim/Joint_FSCIL}}


\end{abstract}

\section{Introduction}

\setlength{\columnsep}{10pt}

\begin{figure}[t]
  \centering
  \vspace{0.5em}
  \begin{subfigure}{.49\columnwidth}
    \centering
    \includegraphics[width=\linewidth]{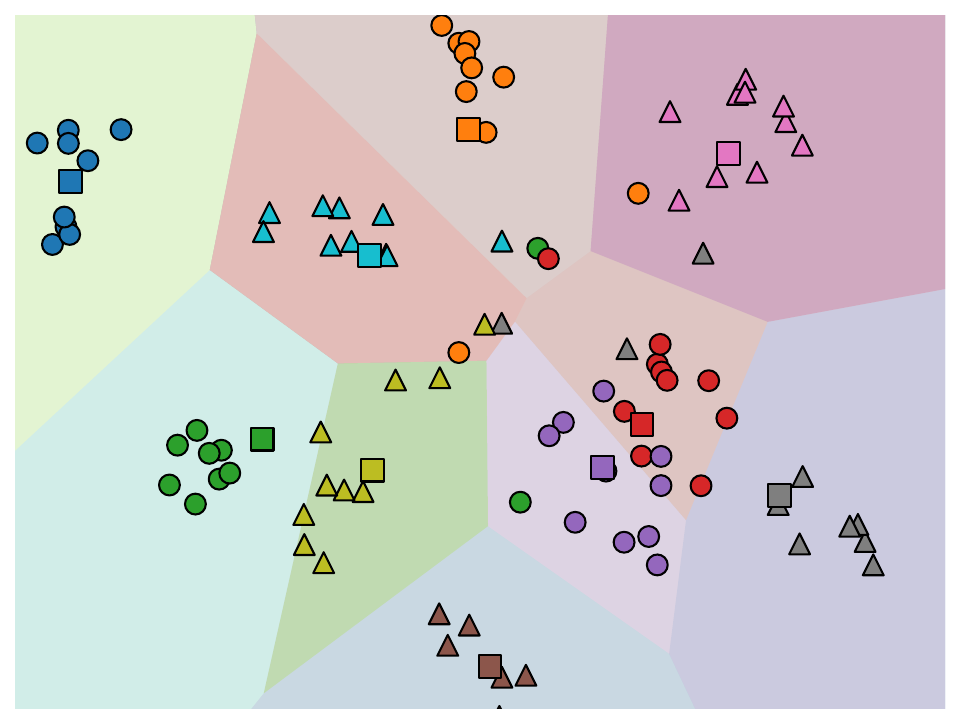}
    \caption{Joint training in CIL setting}
    \label{fig:joint_cil}
  \end{subfigure}%
  \hfill
  \begin{subfigure}{.49\columnwidth}
    \centering
    \includegraphics[width=\linewidth]{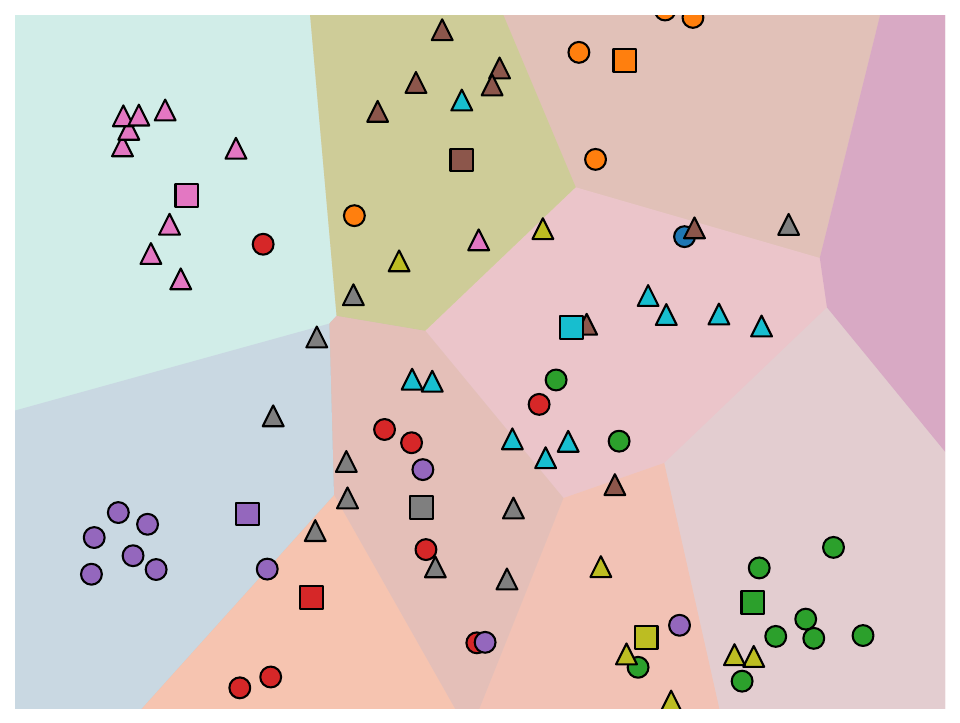}
    \caption{Joint training in FSCIL setting}
    \label{fig:joint_fscil}
  \end{subfigure}
  \caption{Feature space visualization of a joint training model on randomly selected 5 base classes (dots) and 5 incremental classes (triangles) from the CIFAR-100~\citep{cifar100} test set. Class centroids are shown as squares. (a) Joint training in CIL obtains well-clustered features. (b) Joint training in FSCIL results in scattered features.}
  \label{fig:joint_teaser}
\end{figure}

Deep neural networks (DNNs) have achieved remarkable progress in various fields, often matching or even surpassing human capabilities~\citep{jumper2021highly, feng2023learning, jang2025classification}. However, when trained on streaming data, they face the challenge of \textit{catastrophic forgetting} (CF)~\citep{french1999catastrophic, van2022three}, which refers to the loss of previously acquired knowledge when adapting to evolving data distributions. They also struggle with poor \textit{inter-task class separation} (ICS)~\citep{kim2022theoretical, momeni2025achieving}, which leads to ambiguous decision boundaries between previously learned and newly introduced classes. To tackle these issues, class-incremental learning (CIL) has been proposed as a framework that enables models to accommodate new classes over time while maintaining strong performance on all previously observed classes~\citep{yan2021dynamically, zhou2024class}. In this study, we focus on a more practical yet challenging extension of conventional CIL, few-shot class-incremental learning (FSCIL), where new classes emerge with only a few samples~\citep{tao2020few, zhang2025few}. Specifically, the FSCIL task consists of a base session with sufficient training data, followed by multiple incremental sessions where an extremely limited number of samples are provided~\citep{tian2024surveyfewshot}.

Numerous FSCIL approaches have been proposed to address this challenge under the assumption that \textit{previously seen data are no longer accessible in the following incremental sessions}~\citep{zhou2022forward, wang2024few, kim2023warping}. However, in many real-world scenarios such as e-commerce applications or industrial deployments, previously collected datasets often remain available~\citep{prabhu2023computationally, cho2025cost}—albeit possibly large in size or costly to retrain on. This raises a fundamental question: \textit{If access to previous data is allowed, is it better to retrain a model using all accumulated data (i.e., joint training), or to update the model solely based on the newly introduced data}?

The answer to this question is relatively clear in the context of conventional CIL. Given that each incremental session contains a substantial amount of data, joint training is widely regarded as the ideal upper bound~\citep{zhou2024class, momeni2025achieving}. It serves not only as a comparative baseline for evaluating the performance of CIL methods~\citep{van2020brain, lapacz2024exploring, van2022three}, but also as a methodological benchmark that many CIL algorithms try to emulate~\citep{zhou2024class}. For instance, several studies seek to reduce the inductive bias in CIL models by rectifying their classifier weights~\citep{hou2019learning, zhao2020maintaining, ahn2021ss}, output logits~\citep{castro2018end, belouadah2019il2m, wu2019large}, or feature embeddings~\citep{yu2020semantic, zhou2022forward} to align with those of the joint training model. The existence of a well-defined upper bound provides a practical guideline: \textit{when access to previous data is permitted, joint training is preferred for maximizing performance, whereas CIL methods are viable alternatives under constraints in training time or computational resources}.

\begin{figure}[t]
  \centering
  \vspace{2.em}
  \includegraphics[width=1.\linewidth]{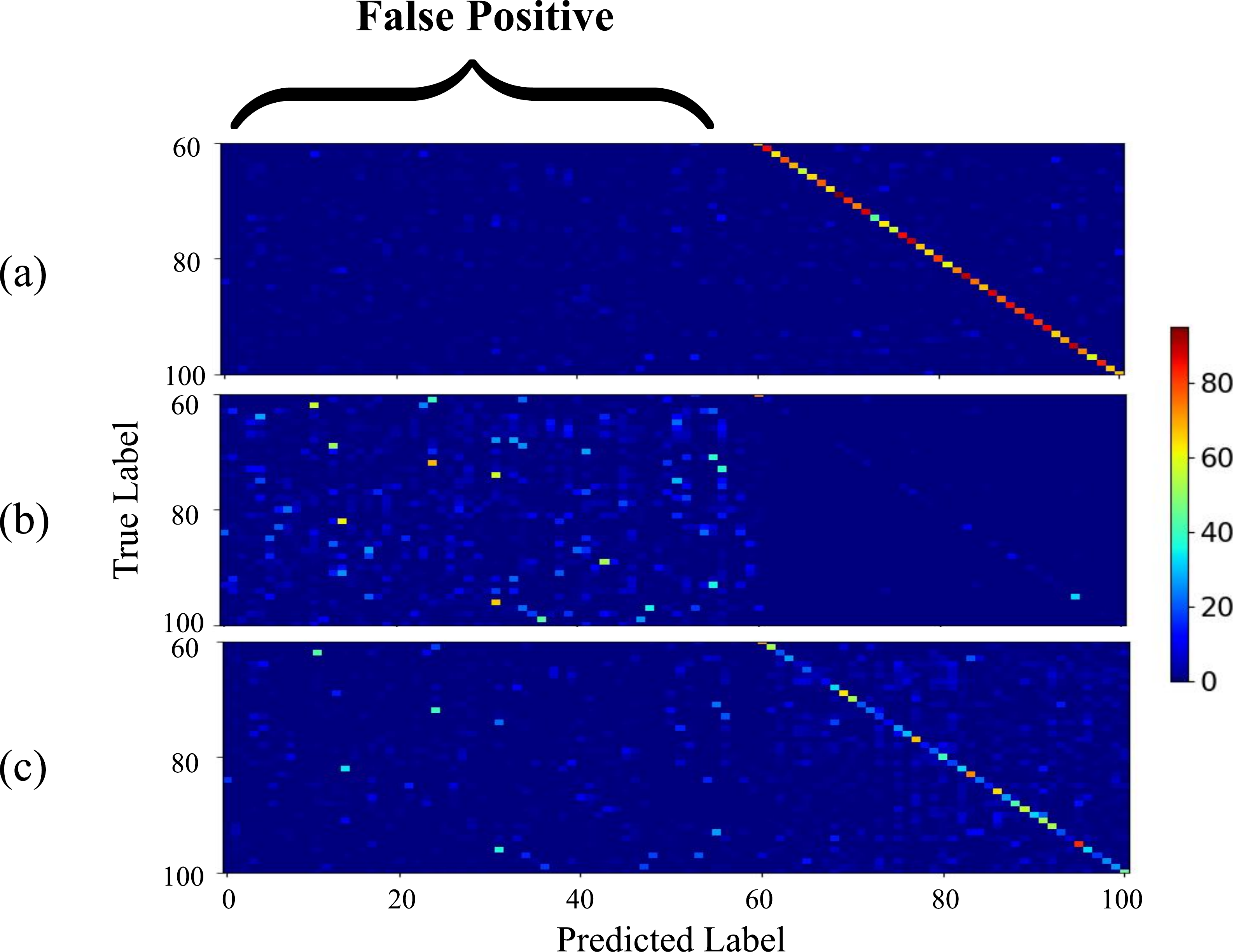}
  \caption{Comparison of confusion matrices on the incremental classes (60-99) of CIFAR-100~\citep{cifar100} test set between standard joint training in (a) CIL setting, (b) FSCIL setting, and (c) imbalance-aware joint training in FSCIL setting. (c) exhibits significantly less false positives for incremental classes than (b).}
  \label{fig:confusion_matrix}
\end{figure}

In contrast to conventional CIL, the severe imbalance between base and incremental classes in FSCIL undermines the effectiveness of joint training as a reliable upper bound. Figure~\ref{fig:joint_teaser} depicts the feature space of a ResNet-20~\cite{he2016deep} joint training model under both CIL and FSCIL settings using t-SNE~\citep{van2008visualizing} and Voronoi diagram~\citep{aurenhammer1991voronoi}. In CIL, incremental class features (triangles) are well separated along decision boundaries (Figure~\ref{fig:joint_cil}). However, in FSCIL, they are scattered and overlapped, failing to form distinct class regions (Figure~\ref{fig:joint_fscil}). The confusion matrices in Figures~\ref{fig:confusion_matrix}\refword{a} and \ref{fig:confusion_matrix}\refword{b} further show that joint training in FSCIL produces notably more false positives for incremental classes than in CIL.

Since joint training proves to be less effective in FSCIL scenarios, it remains unclear whether retraining on the full dataset or incremental learning is preferable. Nevertheless, to the best of our knowledge, no prior work has empirically investigated how to effectively leverage past data in FSCIL settings when it is available.

In this paper, we explore \textit{imbalanced learning} methods as a more realistic joint training benchmark for comparison with FSCIL approaches that do not utilize previous data. Imbalanced learning aims to enhance the representativeness of minority classes, ensuring their contribution to the learning process despite their limited sample size~\citep{imbalance_survey, johnson2019survey, shwartz2023simplifying}. This objective closely aligns with the fundamental assumption of FSCIL, which involves an imbalanced distribution between base and incremental classes~\citep{tian2024surveyfewshot, zhang2025few}.

We categorize eight state-of-the-art imbalanced learning techniques into three taxonomies—resampling-based~\citep{park2022majority, dablain2022deepsmote, ghosh2024class}, reweighting-based~\citep{cao2019learning, cui2019class, ren2020balanced, yang2022inducing}, and optimizer-based~\citep{zhou2023imbsam}—and perform a random search~\citep{mantovani2015effectiveness, bergstra2011algorithms} to identify the optimal combination. We present this combination as a new imbalance-aware joint training benchmark for FSCIL. Figure~\ref{fig:confusion_matrix}\refword{c} demonstrates the effectiveness of the new benchmark in improving the model's ICS. As shown in the confusion matrix, it significantly reduces false positives for incremental classes. This suggests that imbalance-aware joint training offers a more practical and informative reference than conventional joint training for evaluating different approaches in the FSCIL setting.

Based on this insight, we compare its performance with eight state-of-the-art FSCIL methods to provide guidelines for selecting suitable training strategies in few-shot incremental scenarios under varying resource constraints. To ensure fair and consistent comparison, all methods are reimplemented and integrated into a unified framework instead of relying on disparate codebases. Our framework is made public to support reproducibility and provide a transparent pipeline for future FSCIL research.





Our contributions are three-fold:
\begin{itemize}
\item \textit{First}, we initiate a practical discussion on the use of previously observed data in FSCIL. To the best of our knowledge, this is the first empirical study to examine whether retraining or incremental learning is preferable when access to prior data is available in FSCIL settings.

\item \textit{Second}, we investigate the effectiveness of joint training with imbalanced learning strategies in FSCIL scenarios. This serves as a more realistic joint training benchmark for FSCIL that reflects the class imbalance.

\item \textit{Third}, we conduct a comparative analysis of imbalance-aware joint training and state-of-the-art FSCIL methods under varying resource constraints. Our evaluation offers empirical insights into which training strategy is more effective under such conditions.
\end{itemize}


\section{Related Work} 
\subsection{Few-Shot Class-Incremental Learning} 

Few-shot class-incremental learning (FSCIL) addresses the challenge of continually adapting to new classes using only a few samples per class while preserving knowledge of previously learned classes~\cite{tao2020few}. Recent efforts in FSCIL can be broadly categorized into i) \textit{incremental-frozen}, and ii) \textit{fine-tuning} approaches~\cite{roy2024bag}. \textbf{Incremental-frozen approaches} keep the feature extractor fixed during incremental learning, thereby maintaining a stable embedding space for base classes even as novel classes are introduced. While this approach consolidates \textit{stability}—the ability to maintain previous knowledge—it can limit \textit{plasticity}—the ability to learn new patterns—thus motivating the use of various techniques to mitigate this trade-off~\cite{zhou2022forward, song2023learning, yang2023neural, wang2024few, zhou2022few, zhang2021few, tang2025rethinking}. \textbf{Fine-tuning approaches}, on the other hand, update the parameters of the feature extractor partially or entirely in each incremental session, which enhances plasticity at the potential cost of reduced stability~\cite{zhao2021mgsvf, kang2022soft, kalla2022s3c, kim2023warping}. 


In many real-world applications, previous training data often remain accessible even as new data are continuously introduced. However, such scenario has not been considered in existing FSCIL research. This leads to a lack of discussion on proper benchmarks for determining which methods are suitable when prior data is available. In this paper, we explore a new benchmark based on imbalanced learning techniques and compare it against conventional FSCIL methods, providing concrete guidelines for scenarios where past data can be leveraged.

\subsection{Imbalanced Learning} 
Imbalanced learning primarily addresses long-tailed distributions~\cite{imbalance_survey}, where majority classes significantly outnumber minority classes~\cite{zhang2023deep}. Extensive studies have explored strategies to mitigate the resulting model bias, which are commonly categorized into three major approaches: i) \textit{resampling} the training dataset, ii) \textit{reweighting} the objective function, and iii) refining the \textit{optimizer}. \textbf{Resampling-based approaches} include techniques such as CMO~\cite{park2022majority}, which employs a CutMix-based augmentation~\cite{yun2019cutmix} to blend samples from majority and minority classes; DeepSMOTE~\cite{dablain2022deepsmote} which applies GAN-based generation of minority samples; and Ghosh et al.~\cite{ghosh2024class}, which ensure balanced sampling in each training batch. \textbf{Reweighting-based approaches} aim to rebalance gradient signals between majority and minority classes~\cite{cui2019class, ren2020balanced, cao2019learning, yang2022inducing}. \textbf{Optimizer-based approaches} mitigate class imbalance by modifying the optimizer. For example, ImbSAM~\cite{zhou2023imbsam} extends Sharpness-Aware Minimization (SAM)~\cite{foret2020sharpness} by incorporating class-aware weight updates, enhancing generalization under skewed distributions. 

Although these methods are mostly developed for long-tailed datasets, they are not limited to such distributions. Various studies have explored imbalanced learning under different distributional variations. For example, LDAM~\citep{cao2019learning} defines an imbalance ratio and controls the number of samples between major and minor classes accordingly. Buda et al.~\citep{buda2018systematic} introduce step and linear imbalance settings, where the number of samples per class decreases linearly. They also consider extreme cases where all classes except one have very few samples. 

These variations suggest that imbalanced learning hold potential relevance for the FSCIL problem. In this work, we aim to establish a more realistic benchmark for FSCIL that complements the conventional joint training benchmark, by systematically applying and comparing imbalanced learning methods in the FSCIL setting.

\section{Rethinking Joint Training in FSCIL}
\label{sec:joint_training}

To provide a practical and informative guideline for the FSCIL community—particularly in scenarios where access to past data is available—we take a step further by rethinking what constitutes a meaningful benchmark for FSCIL.
From our findings in Figures~\ref{fig:joint_teaser} and ~\ref{fig:confusion_matrix}, we observe that joint training alone cannot serve as a proper benchmark for FSCIL, as it fails to address the inter-task class separation (ICS) problem under severe class imbalance.

To this end, we further explore class imbalance mitigation strategies to establish a more appropriate joint training benchmark for FSCIL, and investigate whether such an approach can serve as a viable standard. We refer to conventional joint training—the baseline method without any modifications—as \textit{standard joint training} throughout this paper, to clearly distinguish it from joint training schemes with imbalance mitigation techniques discussed in Section~\ref{sec:imb_joint_training}.

\subsection{Imbalance-Aware Joint Training in FSCIL}
\label{sec:imb_joint_training}
A fundamental challenge in FSCIL lies in the severe class imbalance between base and incremental classes. Since incremental classes are introduced with only a few samples, the model tends to be biased toward well-represented base classes, resulting in performance degradation. This closely resembles the problem of \textit{imbalanced learning}, which aims to adjust models to learn meaningful representations for underrepresented classes~\citep{imbalance_survey, johnson2019survey, shwartz2023simplifying}.

Motivated by such conceptual similarity, we systematically explore imbalanced learning techniques to develop a reliable joint training benchmark for FSCIL. We first examine how prior studies have addressed class imbalance, and find that many existing works recommend combining independently functioning strategies from different categories of imbalanced learning, such as i) resampling, ii) reweighting, and iii) optimizer-based methods. For instance, Park et al.~\cite{park2022majority} highlight that combining resampling and reweighting techniques leads to significant improvements in the performance of minority classes. In addition, Zhou et al.~\cite{zhou2023imbsam} point out that using only resampling or reweighting without an explicit optimization strategy may cause overfitting or unstable training due to imbalanced gradient updates. These insights suggest that integrating different imbalanced learning methods can yield more stable and robust performance than relying on a single approach.

\begin{figure}[t]
    \centering
    \includegraphics[width=0.99\linewidth]{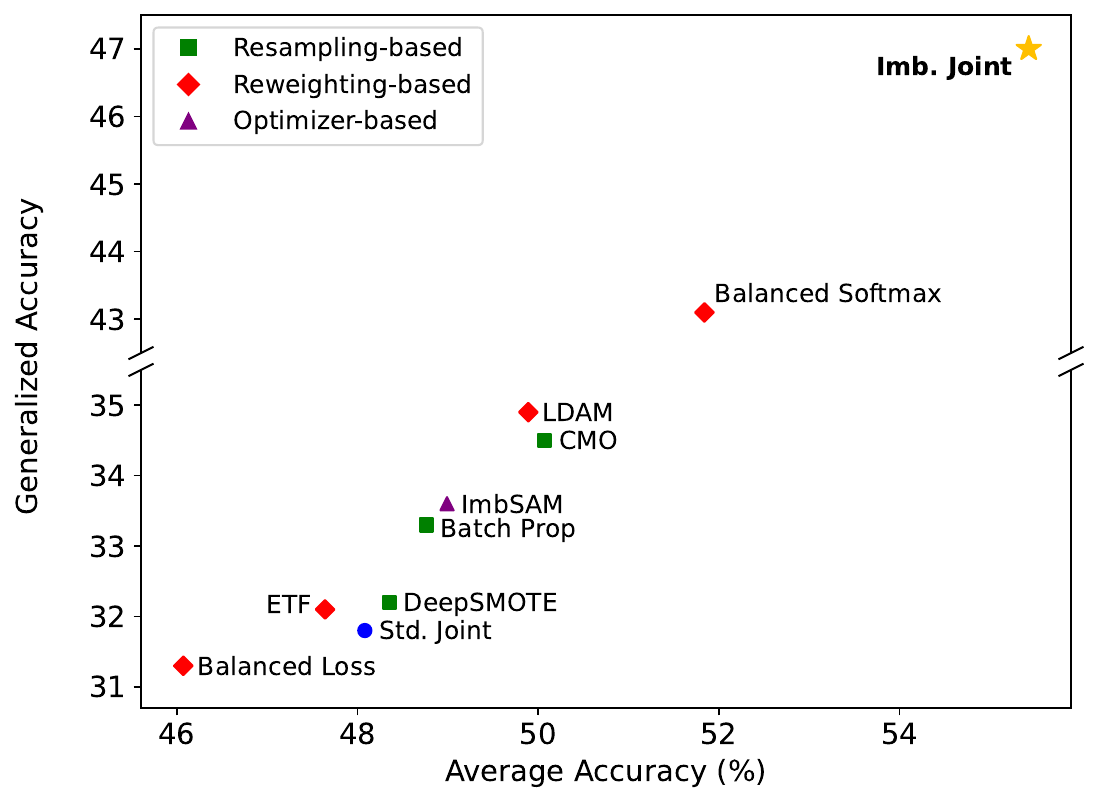}
    \caption{\small
    Performance comparison of imbalance-aware joint training (\textit{Imb. Joint}), standard joint training (\textit{Std. Joint}), and 8 imbalanced learning techniques based on \aacc and \gacc using the CIFAR-100~\citep{cifar100} test set in the last session of a 5-way 5-shot FSCIL setting. Imbalanced learning techniques are presented in three categories: resampling-based, reweighting-based, and optimizer-based. \textit{Imb. Joint} consists of CMO~\cite{park2022majority}, Balanced Softmax~\cite{ren2020balanced}, and ImbSAM~\cite{zhou2023imbsam}, outperforming others by a large margin.
    }
    \label{fig:imbalanced_comp}
\end{figure}

\begin{table}[t]
    \centering
    \small
    \caption{Ablation study of each component in the imbalance-aware joint training benchmark on CIFAR-100~\citep{cifar100} test set. All values are measured in the last session of a 5-way 5-shot FSCIL setting. Each component contributes to the performance improvement.}
    \label{tab:ablation_cifar}
    \resizebox{\linewidth}{!}{
    \begin{tabular}{ccc cc}
        \toprule
        \textit{CMO} & \textit{BalancedSoftmax} & \textit{ImbSAM} & \aacc & \gacc \\
        \midrule
         &  &  & 48.1 & 31.8 \\
        \cmark &  &  & 50.1 & 34.5 \\
        \cmark & \cmark &  & 55.5 & 45.9 \\
        \cmark & \cmark & \cmark & \textbf{55.8} & \textbf{46.8} \\
        \bottomrule
    \end{tabular}
    }
\end{table}

Building on these findings, we combine methods from three categories of imbalanced learning and search for the most effective configuration in the FSCIL setting. Specifically, we classify eight state-of-the-art imbalanced learning methods into resampling~\citep{park2022majority, dablain2022deepsmote, ghosh2024class}, reweighting~\cite{cui2019class, ren2020balanced, cao2019learning, yang2022inducing}, and optimizer-based~\cite{zhou2023imbsam} approaches, and conduct 30 random search trials per method~\cite{mantovani2015effectiveness, bergstra2011algorithms}. We then select the top-performing method from each category and combine them to form the new imbalance-aware joint training benchmark for FSCIL.
Note that, for evaluation, we adopt both average accuracy (\aacc), a standard metric in FSCIL, and generalized average accuracy (\gacc) proposed by Tang et al.~\cite{tang2025rethinking}, which offers a more balanced assessment with explicit emphasis on incremental class performance.

Based on our experiments, we find that combining \textit{CMO} (resampling-based)~\citep{park2022majority}, \textit{Balanced Softmax} (reweighting-based)~\citep{ren2020balanced}, and \textit{ImbSAM} (optimizer-based)~\citep{zhou2023imbsam} achieves the best overall performance. It improves \aacc by 7\%p and \gacc by 15\%p over standard joint training (Table~\ref{tab:ablation_cifar}), and also outperforms all individual imbalanced learning methods (Figure~\ref{fig:imbalanced_comp}). These results indicate that imbalance-aware joint training can serve as a more meaningful reference than standard joint training when developing practical guidelines for real-world FSCIL scenarios. Therefore, we suggest this approach as a new joint training benchmark for FSCIL. A detailed analysis is provided in Section~\ref{sec:analysis_imb}.

\subsection{Analysis of Imbalance-Aware Joint Training}
\label{sec:analysis_imb}

\begin{figure}[t]
    \centering
    \includegraphics[width=0.8\linewidth]{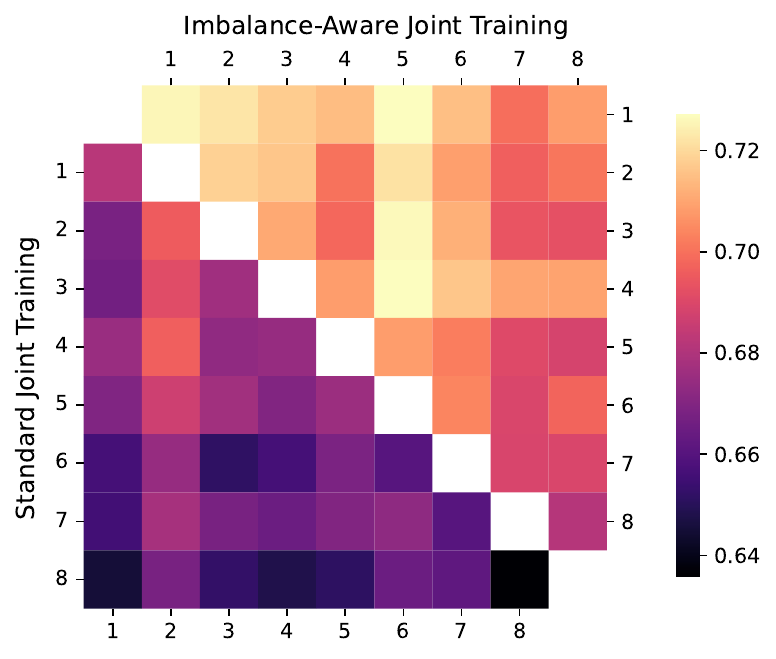}
    \caption{\small
    Feature similarity between joint training models under CIL and FSCIL settings on CIFAR-100~\citep{cifar100} test set based on Centered Kernel Alignment (CKA)~\citep{kornblith2019similarity}. The x- and y-axes represent the incremental sessions. The upper triangular matrix shows the similarity between standard joint training in CIL and imbalance-aware joint training in FSCIL, while the lower triangular matrix presents the similarity between standard joint training in CIL and standard joint training in FSCIL. The brighter coloration in the upper triangle indicates that imbalance-aware joint training in FSCIL yields features more similar to those of standard joint training in CIL than standard joint training in FSCIL does.
    }
    \label{fig:cka}
\end{figure}

\begin{table*}[ht!]
    \small
    \centering
    \setlength{\tabcolsep}{6pt}
    \caption{Comparison of base session training setups of 8 FSCIL methods~\cite{zhang2021few, kalla2022s3c, kim2023warping, zhou2022forward, wang2024few, song2023learning, zhou2022few, tang2025rethinking}.}
    \begin{tabular}{lcccccccc}
        \toprule
        & CEC & S3C
        & WaRP & FACT & TEEN & SAVC & LIMIT & Yourself \\
        \midrule
        \textbf{P1}: Exposure of test set during training           
        & \cmark & \cmark & \xmark & \xmark & \cmark & \cmark & \cmark & \xmark \\
        
        
        
        \textbf{P2}: Unfair usage of pre-trained encoders  
        & \xmark & \xmark & \xmark & \xmark & \xmark & \xmark & \xmark & \cmark \\
        \bottomrule
    \end{tabular}
    \label{tab:comparison}
\end{table*}

\begin{table*}[ht!]
    \centering
    \fontsize{8}{8}\selectfont
    \setlength{\tabcolsep}{7.5pt}  
    \renewcommand{\arraystretch}{1.1}  
    \caption{Comparison of methods across multiple sessions on CIFAR-100~\citep{cifar100}. \textbf{S0} presents the base session and \textbf{S1-S8} denote incremental sessions. The best and second-best results are \textbf{bolded} and \underline{underlined}. All methods are reproduced within our unified codebase. \textit{Std. Joint} and \textit{Imb. Joint} denote standard joint training and imbalance-aware joint training, respectively.}
    \label{tab:cifar100_comparison}
    \begin{tabular}{@{}p{1.8cm}|c|ccccccccc|cc|cc@{}}
    \toprule
    \multicolumn{1}{c|}{} & {\textbf{}} & \multicolumn{9}{c|}{\aacc \textbf{in each session (\%)}} & \multicolumn{2}{c|}{\aacc \textbf{S8}} \\
    \textbf{Method} & 
    \textbf{Architecture} &
    \textbf{S0} & 
    \textbf{S1} & 
    \textbf{S2} & 
    \textbf{S3} & 
    \textbf{S4} & 
    \textbf{S5} & 
    \textbf{S6} & 
    \textbf{S7} & 
    \textbf{S8} & 
    \textbf{Base} & 
    \textbf{Inc.} & 
    \aacc & 
    \gacc \\
    \midrule \midrule
    Std. Joint & \multirow{2}{*}{ResNet-20}
    & 77.5 & 73.1 & 68.1 & 63.2 & 59.7 & 56.3 & 53.1 & 51.2 & 48.1
    & \textbf{78.8} 
    & 1.9
    & 61.1
    & 46.9 \\
    Imb. Joint & 
    & 78.4 & 75.0 & 71.6 & 66.9 & 64.0 & 62.5 & 60.0 & 59.1 & \textbf{55.3}
    & 70.5 
    & \underline{32.5}
    & \textbf{65.9}
    & \textbf{58.0} \\
    \midrule
    CEC~\citep{zhang2021few} & \multirow{7}{*}{ResNet-20}
    & 76.4 & 57.0 & 53.1 & 50.0 & 47.6 & 45.3 & 43.7 & 41.7 & 39.7
    & 50.5 
    & 23.5
    & 50.5
    & 45.9 \\
    FACT~\citep{zhou2022forward} &
    & 68.2 & 63.0 & 58.4 & 54.7 & 51.6 & 48.9 & 46.6 & 44.3 & 42.4
    & 62.5 
    & 12.1
    & 53.1
    & 43.6 \\
    TEEN~\citep{wang2024few} &
    & 67.0 & 62.3 & 58.1 & 54.5 & 51.2 & 48.7 & 46.1 & 43.8 & 41.7
    & 63.7 
    & 8.8
    & 52.6
    & 42.4 \\
    S3C~\citep{kalla2022s3c} &
    & 56.6 & 54.7 & 52.4 & 49.2 & 47.5 & 46.1 & 44.7 & 43.5 & 41.3
    & 47.4 
    & 32.1
    & 48.4
    & 45.5 \\
    WaRP~\citep{kim2023warping} &
    & 70.0 & 66.2 & 62.5 & 58.5 & 55.4 & 52.7 & 51.2 & 49.1 & 47.1
    & 64.2 
    & 21.6
    & 57.0
    & 49.2 \\
    SAVC~\citep{song2023learning} &
    & 80.5 & 76.0 & 71.5 & 67.4 & 64.1 & 61.3 & 59.2 & 57.0 & \underline{54.7}
    & \underline{76.5} 
    & 22.2
    & \underline{65.8}
    & \underline{55.6} \\
    LIMIT~\citep{zhou2022few} &
    & 74.1 & 69.9 & 66.2 & 62.2 & 59.1 & 56.3 & 54.2 & 52.1 & 49.7
    & 68.9 
    & 21.1
    & 60.4
    & 51.2 \\ 
    YourSelf~\citep{tang2025rethinking}
    & DeiT-S
    & 71.6 & 67.2 & 64.1 & 60.4 & 57.4 & 54.6 & 52.8 & 51.2 & 48.5
    & 56.0 
    & \textbf{37.3}
    & 58.7
    & 54.6 \\
    \bottomrule
\end{tabular}

\end{table*}

To evaluate the individual effectiveness of different types of imbalanced learning techniques, we conduct ablation experiments on the imbalance-aware joint training benchmark using the CIFAR-100 test set. As demonstrated in Table~\ref{tab:ablation_cifar}, the standard joint training model achieves 48.1\% on \aacc and 31.8\% on \gacc.
We then independently apply representative methods from each category—CMO, Balanced Softmax, and ImbSAM. Each method yields performance improvements, with the best results reaching 55.8\% on \aacc and 46.8\% on \gacc.
These findings indicate that each technique independently enhances performance, confirming that imbalanced learning methods from different categories provide complementary benefits.

Furthermore, to evaluate how effectively the imbalance-aware joint training approach addresses the ICS problem in FSCIL, we use Centered Kernel Alignment (CKA)~\citep{kornblith2019similarity} to measure the similarity of network representations between joint training models in CIL and FSCIL settings (Figure~\ref{fig:cka}). The upper triangular matrix shows the similarities between imbalance-aware joint training in FSCIL and standard joint training in CIL, while the lower one represents the similarities between standard joint training in FSCIL and CIL. As indicated by the brighter coloration, the upper triangle exhibits consistently higher similarity across all sessions. This observation suggests that imbalance-aware joint training in FSCIL produces representations more closely aligned with those of standard joint training in CIL than standard joint training in FSCIL does.

\begin{table*}[t]
    \centering
    \fontsize{8}{8}\selectfont
    \setlength{\tabcolsep}{7.5pt}  
    \renewcommand{\arraystretch}{1.1}  
    \caption{Comparison of methods across multiple sessions on \emph{mini}ImageNet~\citep{miniimagenet}. \textbf{S0} represents the base session and \textbf{S1-S8} correspond to incremental sessions. The best and second-best results are \textbf{bolded} and \underline{underlined}. All methods are reproduced within our unified codebase. \textit{Std. Joint} and \textit{Imb. Joint} denote standard joint training and imbalance-aware joint training, respectively.}
    \label{tab:mini_comparison}
    \begin{tabular}{@{}p{1.8cm}|c|ccccccccc|cc|cc@{}}
    \toprule
    \multicolumn{1}{c|}{} & \multirow{2}{*}{\textbf{Architecture}} & \multicolumn{9}{c|}{\aacc \textbf{in each session (\%)}} & \multicolumn{2}{c|}{\aacc \textbf{S8}} \\
    \textbf{Method} & 
    & \textbf{S0} & 
    \textbf{S1} & 
    \textbf{S2} & 
    \textbf{S3} & 
    \textbf{S4} & 
    \textbf{S5} & 
    \textbf{S6} & 
    \textbf{S7} & 
    \textbf{S8} & 
    \textbf{Base} & 
    \textbf{Inc.} & 
    \aacc & 
    \gacc \\
    \midrule \midrule
    Std. Joint & \multirow{2}{*}{ResNet-18}
    & 71.0 & 66.7 & 62.1 & 58.5 & 54.8 & 51.5 & 48.7 & 46.3 & 44.2
    & \underline{72.3} 
    & 2.1
    & 56.0
    & 43.0 \\
    Imb. Joint & 
    & 71.0 & 69.4 & 65.3 & 62.9 & 59.0 & 57.1 & 55.1 & 53.6 & \underline{51.7}
    & 66.7 
    & \underline{29.1}
    & \underline{60.6}
    & \underline{53.5} \\
    \midrule
    CEC~\citep{zhang2021few} & \multirow{7}{*}{ResNet-18}
    & 70.9 & 65.0 & 61.1 & 58.1 & 55.5 & 52.7 & 50.1 & 48.2 & 46.7
    & 65.2 
    & 18.9
    & 56.5
    & 47.9 \\
    FACT~\citep{zhou2022forward} &
    & 69.5 & 64.7 & 60.4 & 57.0 & 53.8 & 50.8 & 48.0 & 46.0 & 44.1
    & 66.8 
    & 9.9
    & 54.9
    & 44.4 \\
    TEEN~\citep{wang2024few} &
    & 64.9 & 60.7 & 56.9 & 54.4 & 52.0 & 49.4 & 47.0 & 45.2 & 43.8
    & 58.0 
    & 22.4
    & 52.7
    & 45.8 \\
    S3C~\citep{kalla2022s3c} &
    & 57.7 & 53.9 & 51.1 & 49.0 & 47.6 & 45.1 & 42.8 & 41.4 & 40.7
    & 51.4 
    & 24.9
    & 47.7
    & 42.3 \\
    WaRP~\citep{kim2023warping} &
    & 71.5 & 66.7 & 63.0 & 60.2 & 57.7 & 55.1 & 52.5 & 50.9 & 49.7
    & 65.9 
    & 25.4
    & 58.6
    & 50.8 \\
    SAVC~\citep{song2023learning} &
    & 80.0 & 75.4 & 71.2 & 67.5 & 64.5 & 61.1 & 58.1 & 56.0 & \textbf{54.1}
    & \textbf{76.2} 
    & 21.1
    & \textbf{65.3}
    & \textbf{55.6} \\
    LIMIT~\citep{zhou2022few} &
    & 72.9 & 66.4 & 62.3 & 59.0 & 56.0 & 53.3 & 50.5 & 48.6 & 47.1
    & 64.2 
    & 21.6
    & 57.3
    & 49.2 \\
    YourSelf~\citep{tang2025rethinking} & DeiT-S
    & 71.8 & 66.0 & 62.3 & 59.4 & 57.4 & 54.5 & 52.0 & 50.5 & 49.4
    & 60.9 
    & \textbf{32.2}
    & 58.2
    & 52.6 \\
    \bottomrule
\end{tabular}

\end{table*}

\begin{table*}[t]
    \centering
    \fontsize{8}{8}\selectfont
    \setlength{\tabcolsep}{5.7pt}  
    \renewcommand{\arraystretch}{1.1}  
    \caption{Comparison of methods across multiple sessions on CUB-200~\citep{cub200}. \textbf{S0} presents the base session and \textbf{S1-S10} denote incremental sessions. The best and second-best results are \textbf{bolded} and \underline{underlined}. All methods are reproduced within our unified codebase. \textit{Std. Joint} and \textit{Imb. Joint} denote standard joint training and imbalance-aware joint training, respectively.}
    \label{tab:cub200_comparison}
    \begin{tabular}{@{}p{1.8cm}|c|ccccccccccc|cc|cc@{}}
    \toprule
    \multicolumn{1}{c|}{} & \multirow{2}{*}{\textbf{Architecture}} & \multicolumn{11}{c|}{\aacc \textbf{in each session (\%)}} & \multicolumn{2}{c|}{\aacc \textbf{S10}} \\
    \textbf{Method} & 
    & \textbf{S0} & 
    \textbf{S1} & 
    \textbf{S2} & 
    \textbf{S3} & 
    \textbf{S4} & 
    \textbf{S5} & 
    \textbf{S6} & 
    \textbf{S7} & 
    \textbf{S8} & 
    \textbf{S9} & 
    \textbf{S10} & 
    \textbf{Base} & 
    \textbf{Inc.} & 
    \aacc & 
    \gacc \\
    \midrule \midrule
    Std. Joint & \multirow{2}{*}{ResNet-18}
    & 76.8 & 73.0 & 69.4 & 66.5 & 64.8 & 63.8 & 60.5 & 59.7 & 60.6 & 59.5 & 58.3
    & \textbf{77.4} 
    & 40.1
    & 62.6
    & 57.0 \\
    Imb. Joint &  
    & 77.5 & 74.9 & 71.9 & 68.1 & 67.9 & 65.3 & 63.6 & 62.6 & 62.5 & 62.3 & \underline{61.9}
    & 73.2 
    & \underline{51.4}
    & \underline{65.1}
    & \underline{62.8} \\
    \midrule
    CEC~\citep{zhang2021few} & \multirow{7}{*}{ResNet-18}
    & 72.6 & 68.2 & 63.3 & 58.0 & 57.4 & 53.1 & 51.0 & 49.3 & 46.5 & 46.2 & 44.6
    & 66.2 
    & 24.1
    & 52.2
    & 48.1 \\
    FACT~\citep{zhou2022forward} &
    & 77.8 & 73.7 & 70.0 & 65.6 & 64.3 & 61.2 & 59.8 & 58.9 & 57.6 & 56.5 & 55.2
    & 73.0 
    & 38.1
    & 61.0
    & 57.1 \\
    TEEN~\citep{wang2024few} &
    & 78.6 & 73.7 & 69.8 & 64.8 & 64.0 & 60.6 & 59.7 & 58.8 & 57.6 & 56.0 & 54.9
    & 70.5 
    & 40.0
    & 60.7
    & 57.8 \\
    S3C~\citep{kalla2022s3c} &
    & 62.1 & 60.6 & 57.8 & 54.3 & 55.3 & 52.5 & 51.9 & 51.0 & 50.7 & 50.4 & 49.9
    & 54.5 
    & 44.1
    & 52.7
    & 52.1 \\
    WaRP~\citep{kim2023warping} &
    & 77.0 & 73.4 & 70.0 & 66.0 & 64.8 & 61.8 & 60.7 & 57.9 & 58.2 & 57.3 & 56.2
    & 72.3 
    & 40.9
    & 61.4
    & 57.9 \\
    SAVC~\citep{song2023learning} &
    & 78.0 & 75.0 & 71.9 & 67.6 & 67.1 & 64.4 & 63.8 & 61.9 & 61.0 & 60.5 & 59.9
    & \underline{75.1} 
    & 45.4
    & 64.3
    & 60.4 \\
    LIMIT~\citep{zhou2022few} & 
    & 67.3 & 63.1 & 58.7 & 54.4 & 53.3 & 49.4 & 46.9 & 44.8 & 43.4 & 42.3 & 40.3
    & 58.1 
    & 23.3
    & 48.2
    & 44.2 \\
    YourSelf~\citep{tang2025rethinking} & DeiT-S
    & 80.8 & 77.8 & 74.7 & 72.0 & 68.9 & 65.5 & 64.6 & 64.1 & 62.5 & 63.1 & \textbf{62.5}
    & 73.4 
    & \textbf{52.3}
    & \textbf{66.4}
    & \textbf{64.3} \\
    \bottomrule
\end{tabular}

\end{table*}

\section{Towards a Practical Guideline for FSCIL}
\label{sec:experiments}

\subsection{Experimental Setup}
\subsubsection{General Settings}
\myparagraph{Dataset.} Following prior works, we conduct experiments on CIFAR-100~\cite{cifar100}, \emph{mini}ImageNet~\cite{miniimagenet}, and CUB-200~\cite{cub200} datasets using the data splits in Tao et al.~\cite{tao2020few}. For CIFAR-100 and \emph{mini}ImageNet, 60 classes are allocated to the base session, and each of the 8 incremental sessions contains 5 classes. For CUB-200, 100 classes are used for the base session, followed by 10 incremental sessions with 10 classes each. All datasets are evaluated under the 5-shot setting.

\myparagraph{Evaluation Metrics.} Following common practice in FSCIL research~\cite{tao2020few, zhang2021few}, we use average accuracy (\aacc) as our primary evaluation metric. Additionally, unlike prior works that report only the overall average accuracy, we separately report the average accuracy of base classes and incremental classes. We also adopt generalized average accuracy (\gacc) proposed by Tang et al.~\cite{tang2025rethinking}, which balances the evaluation of base and incremental classes using a tunable parameter $\alpha$ that controls their respective weights.

\myparagraph{Implementation Details.} As the backbone feature extractor, all methods except for YourSelf~\citep{tang2025rethinking} utilize ResNet-20~\citep{resnet} for CIFAR-100, and ResNet-18 for \emph{mini}ImageNet and CUB-200. YourSelf employs DeiT-S~\cite{deit}, a ViT-based architecture, across all three datasets.
All our experiments are conducted on a single NVIDIA A5000 GPU. To ensure a consistent environment, we incorporate all methods into a unified codebase. All reimplementations are based on publicly available GitHub repositories.

\subsubsection{A Standardized Evaluation Protocol for FSCIL}
\label{sec:fair_comp}
In this paper, we evaluate existing FSCIL methods and joint training approaches after resolving inconsistencies in their experimental setups. Although most methods follow a similar training pipeline, subtle but unfair differences undermine the reliability of performance comparisons. To address this, we identify two major inconsistencies as shown in Table~\ref{tab:comparison}, and standardize them to enable a unified and fair comparison of eight FSCIL methods and joint training.

\myparagraph{Exposure of test set during training (P1).} A major issue in prior FSCIL research is the use of the test set as a validation set. Many methods select the best-performing epoch in the base session using the test set. Some methods even use the test set from the last incremental session—which covers the entire label space of the dataset—for hyperparameter tuning~\cite{zou2022margin, zhang2021few}. This leakage can lead to overfitting to the test data and result in unreliable evaluations of generalization performance~\cite{blum2015ladder}. WaRP~\cite{kim2023warping} acknowledges this issue and avoids the usage of the test set by selecting the checkpoint from the final epoch instead.

Such inconsistent usage of the test set prevents fair comparisons across methods. To resolve this, we create a new validation set by splitting the original training set in a 9:1 ratio. In addition, for methods that retrain the entire model during incremental sessions~\cite{tang2025rethinking}, we standardize the evaluation by using model weights from the final epoch, since reserving a separate validation set is impractical due to the limited size of incremental data.

\myparagraph{Unfair usage of pre-trained encoders (P2).} Another issue is that some FSCIL methods leverage additional information from pre-trained encoders. For example, YourSelf~\cite{tang2025rethinking} employs knowledge distillation from a CNN-based state-of-the-art teacher model~\cite{yang2023neural} to accelerate the convergence of a ViT-based encoder. This teacher model requires prior knowledge of the total number of classes, which may compromise the fairness of the comparison. To ensure consistency, we modify YourSelf to perform knowledge distillation only from a model trained under our standardized evaluation protocol, without such additional information. Likewise, we exclude Park et al.~\cite{park2024pre} from our comparison, as it uses large-scale pre-trained encoders like CLIP that already demonstrate strong zero-shot classification performance.

\begin{figure*}[t]
    \centering
    \subfloat[Average Accuracy (\aacc)]{
        \includegraphics[width=0.48\linewidth]{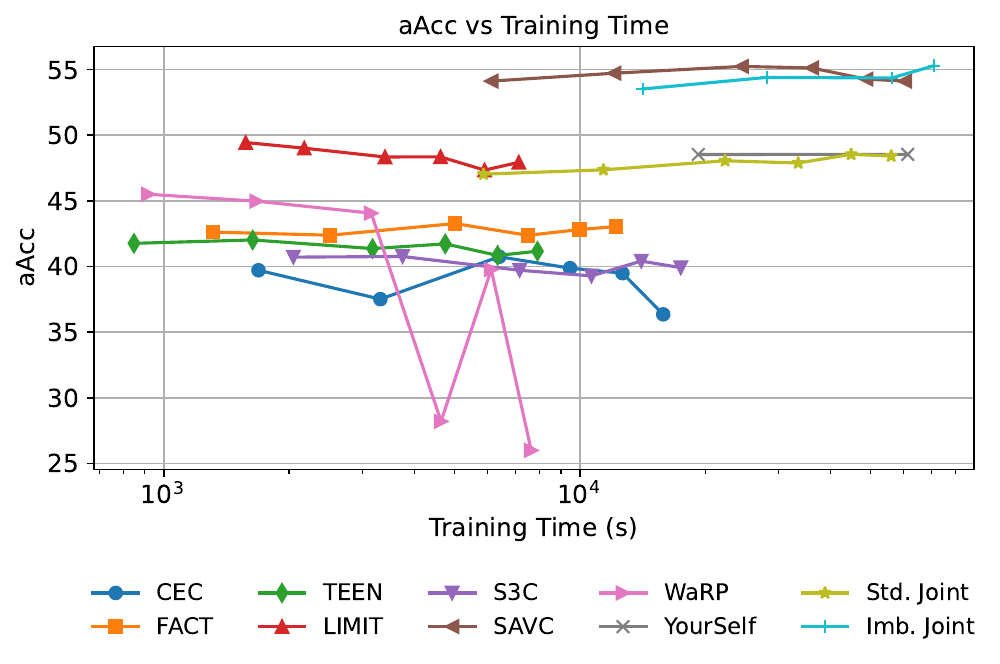}
        \label{fig:metric_acc}
    }
    \hfill
    \subfloat[Generalized Average Accuracy (\gacc)]{
        \includegraphics[width=0.48\linewidth]{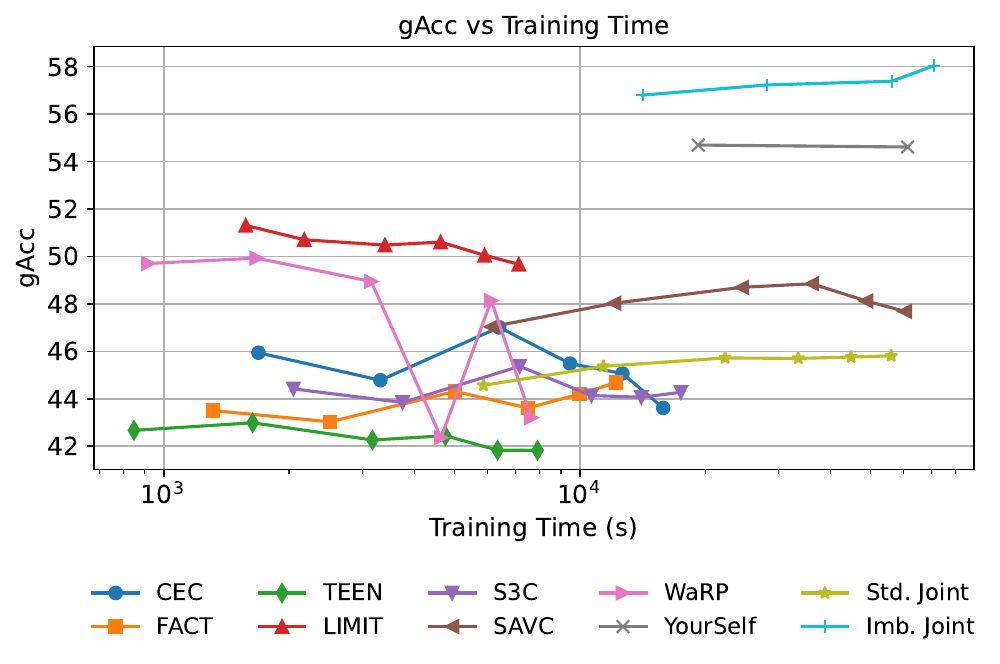}
        \label{fig:metric_gacc}
    }
    \caption{\small
    Performance comparison across different training times on CIFAR-100~\citep{cifar100} test set. Training times are presented on a log scale. \textit{Std. Joint} and \textit{Imb. Joint} denote standard joint training and imbalance-aware joint training, respectively.
    }
    \label{fig:acc_gacc}
\end{figure*}

\subsection{Comparison of FSCIL and Joint Training}

In this section, we conduct a comprehensive comparison of existing FSCIL methods and joint training approaches (\textit{i.e.}, standard joint training and imbalance-aware joint training) across three datasets. We then provide an in-depth analysis and discussion on which approach is most suitable, depending on the availability of previously used training data.

First, we present the experimental results on CIFAR-100 in Table~\ref{tab:cifar100_comparison}. The results show that the standard joint training approach achieves strong performance only on the base classes, while its incremental accuracy is extremely low at 1.9\%. In contrast, when incorporating imbalanced learning techniques to joint training, we observe a significant improvement in \iacc (32.5\%), which in turn leads to improvements in \aacc and \gacc.

Notably, this imbalance-aware joint training approach also outperforms existing FSCIL methods in both \aacc and \gacc. While YourSelf yields the highest incremental class accuracy among all methods, it shows relatively low performance on the base classes, resulting in lower overall performance compared to imbalance-aware joint training. These findings suggest that, when the previous training dataset is accessible, imbalance-aware joint training may be a more effective strategy than FSCIL methods.

However, unlike in CIFAR-100, FSCIL methods outperform imbalance-aware joint training on the other datasets. On \emph{mini}ImageNet, for example, the FSCIL method SAVC achieves the best overall results with an \aacc of 65.3\% and a \gacc of 55.6\%, surpassing imbalance-aware joint training by 4.7\%p and 2.1\%p, respectively (Table~\ref{tab:mini_comparison}). Similarly, on CUB-200, YourSelf demonstrates a \gacc of 64.3\%, outperforming the joint training approach (62.8\%) as shown in Table~\ref{tab:cub200_comparison}. The fact that FSCIL methods—without access to previous training data—can perform better than the imbalance-aware joint training approach that leverages such data challenges the conventional belief that more information necessarily leads to better performance.

\myparagraph{Discussion.} The results in this section suggest that, despite having access to additional data, current imbalanced learning techniques may perform worse than FSCIL approaches under extreme data distributions—particularly when only a few classes have very limited samples. This highlights the need for future research on imbalanced learning approaches that can better handle such challenging scenarios. Since FSCIL methods have shown strong performance in these cases, their strategies could be effectively adapted for imbalanced learning. In particular, since previous training datasets are typically accessible in imbalanced learning, techniques in FSCIL that simulate past data can be replaced with actual use of previous dataset for direct application to imbalanced learning problems. For example, while YourSelf stores only the distribution of past data, this strategy can be extended to directly utilize the full previous training dataset.

\subsection{Resource-Aware Comparison}

In this section, we analyze each method based on its training time to provide practical guidelines for users on which approach to adopt depending on available training resources. Training efficiency is assessed using two metrics (\aacc and \gacc). To account for varying training times, we evaluate the performance of each model at [$100$, $200$, $400$, $600$, $800$, $1000$] epochs. However, due to their longer training times, the imbalance-aware joint training and YourSelf methods are evaluated only at [$100$, $200$, $400$, $500$] epochs and [$100$, $200$] epochs, respectively.

Figure~\ref{fig:metric_acc} shows the \aacc values for each method over training time. We observe that FSCIL methods with longer training durations, such as SAVC and YourSelf, generally achieve higher \aacc than those with shorter training times. Figure~\ref{fig:metric_gacc} presents the \gacc values for each method, which reveal a different trend. Unlike the results in \aacc, methods with longer training times, including SAVC and standard joint training, achieve lower \gacc. This suggests that they maintain strong base class performance but struggle with incremental class learning. In contrast, the imbalance-aware joint training approach consistently records the highest \gacc across all epochs, outperforming FSCIL methods regardless of training duration.

Interestingly, some methods with shorter training times, such as LIMIT and TEEN, show a declining trend in \gacc as training progresses. This pattern suggests that prolonged training on the base session can lead to overfitting to base classes. Conversely, FACT exhibits the opposite tendency, with \gacc improving over time. This indicates that FACT retains greater flexibility during training and benefits from longer training durations.

These results suggest that when users possess sufficient computational resources and access to the previous training dataset, the imbalance-aware joint training method can be a viable choice. However, in cases where resources are sufficient but access to prior data is restricted, SAVC or YourSelf are strong alternatives, despite their longer training times. When both computational resources and access to previous data are limited, LIMIT offers a better balance between efficiency and overall performance.

\section{Conclusion}

Few-shot class-incremental learning (FSCIL) is particularly challenging, as models must continually accommodate new classes with only a few samples per class. In this paper, we highlight a practical but relatively underexplored problem in the FSCIL literature: \textit{the lack of an established benchmark for evaluating whether leveraging previously learned data, when available, is beneficial in the FSCIL setting}. We point out that standard joint training, which serves as the upper bound in conventional CIL, is unsuitable as a benchmark in FSCIL due to its instability under imbalanced data distributions. To address this issue, we explore imbalanced learning techniques that enhance the performance of joint training in FSCIL and suggest a new joint training benchmark. We then conduct extensive experiments to compare this imbalance-aware joint training benchmark with state-of-the-art FSCIL methods. Based on these comparisons, we offer practical guidelines for determining whether utilizing past data is beneficial in FSCIL scenarios.

\myparagraph{Limitations.} We acknowledge that we are unable to reproduce a broader range of recent FSCIL methods and therefore cannot include them in our comparisons. Additionally, while our experiments allow for a performance comparison between FSCIL and joint training, we cannot provide a detailed analysis of why certain methods outperform others due to the limited number of datasets used in the evaluation. Future work focuses on covering a more diverse set of FSCIL approaches and conducting experiments on a broader range of real-world datasets, thereby providing more comprehensive and practical insights.

\section*{Acknowledgement}
This work was supported by the Institute of Information \& communications Technology Planning \& Evaluation (IITP) grant funded by the Korea government(MSIT) (RS-2022-00143911, AI Excellence Global Innovative Leader Education Program).

{
    \small
    \bibliographystyle{ieeenat_fullname}
    \bibliography{ref}
}

\clearpage

\appendix
\renewcommand\thefigure{\thesection\arabic{figure}} 
\setcounter{figure}{0}
\renewcommand{\thetable}{A\arabic{table}}
\setcounter{table}{0}
\section*{Appendix}

\section{Additional Analysis}

\begin{figure}[ht]
    \centering
    \includegraphics[width=0.99\linewidth]{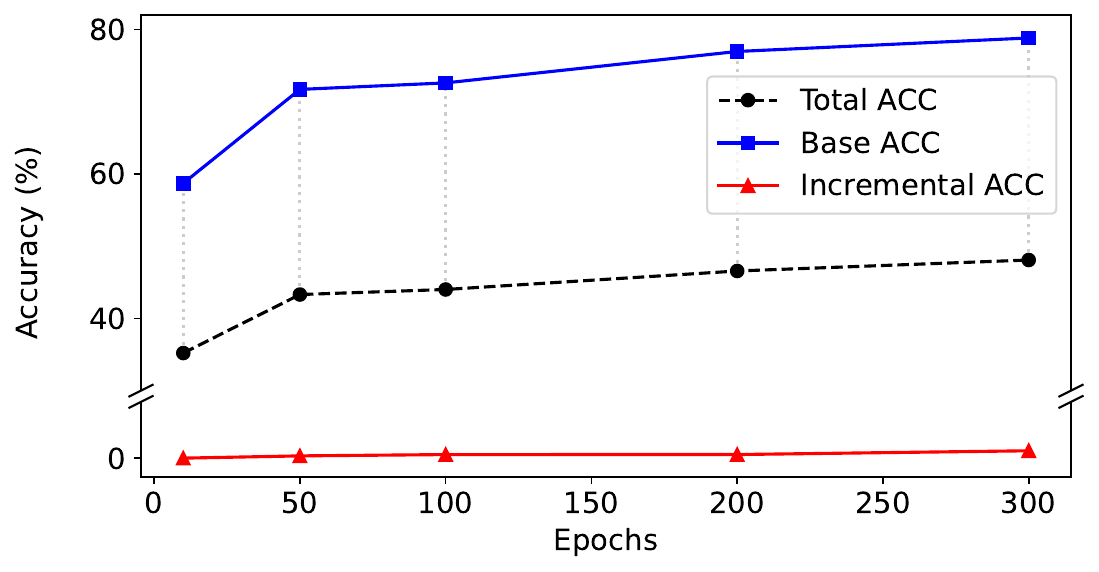}
    \vspace{-0.5em}
    \caption{\small
    Average accuracies of base and incremental classes in standard joint training on the CIFAR-100 test set by the number of base session training epochs. All results are reported from the last incremental session.
    }
    \label{fig:base_incr_joint}
\end{figure}

\myparagraph{Performance Bias in Standard Joint Training.}
Figure~\ref{fig:base_incr_joint} illustrates the average accuracies of base and incremental classes in the FSCIL setting for standard joint training. We observe that standard joint training achieves near-zero accuracy on incremental classes, and its overall performance is heavily biased toward base classes. Such gap between base and incremental accuracies grows linearly as the number of training epochs in the base session increases.

\begin{table}[ht]
\centering
\small
\captionof{table}{\small Average false positive (FP) rate and false negative (FN) rate of standard joint training on CIFAR-100 test set under both the CIL and FSCIL settings.} 
\vspace{-1em}
\begin{tabular}{lcc}\\\toprule  
  Method & FP rate  & FN rate \\ 
  \midrule \midrule
  Standard Joint Training (CIL) & 0.254 & 0.255 \\
  Standard Joint Training (FSCIL) & 0.488 & 0.521 \\ 
  \bottomrule
\end{tabular}
\label{tab:false_positive_rate}
\end{table} 

\myparagraph{Inter-Task Class Separation (ICS) in Joint Training.}
To highlight the ICS problem in standard joint training under the FSCIL setting, we quantitatively evaluate the quality of decision boundary formation using false positive (FP) and false negative (FN) rates. As shown in Table~\ref{tab:false_positive_rate}, the joint training model exhibits higher FP and FN rates in the FSCIL setting than in the CIL setting on average. This shows that the ICS problem is more severe in standard joint training under the FSCIL setting compared to the CIL setting.

\vspace{.5em}

\begin{figure}[ht]
    \centering
    \vspace{2.0em}
    \includegraphics[width=1.\linewidth]{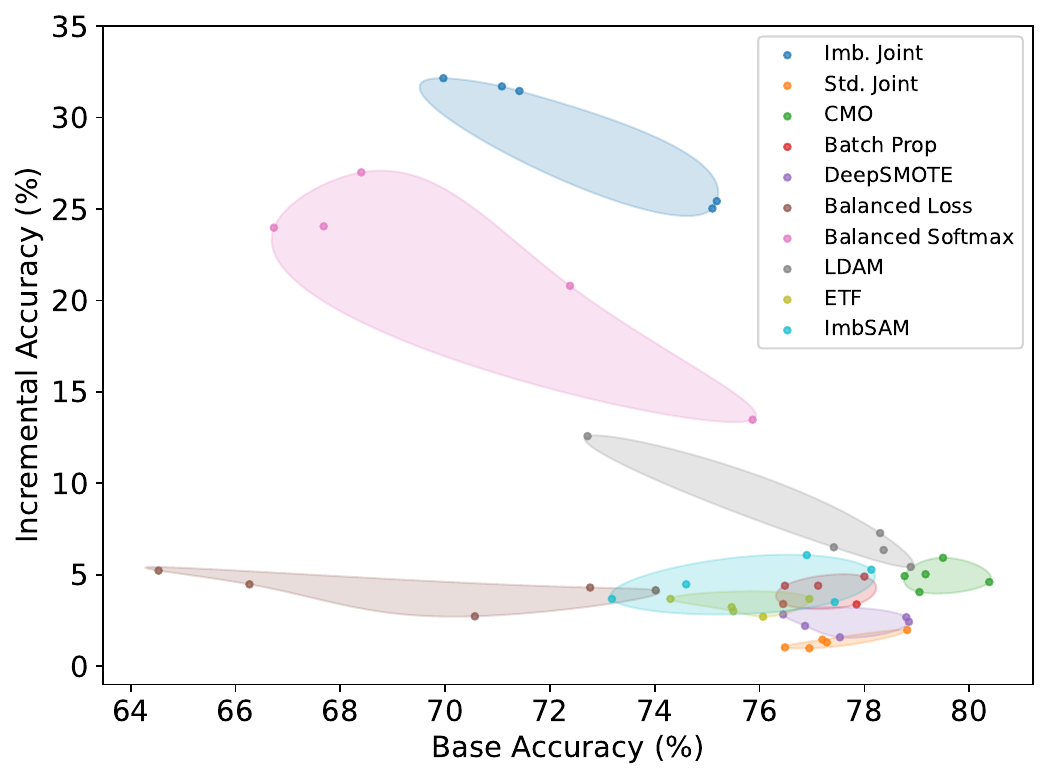}
    \caption{\small
    Base and incremental class performance comparison of imbalance-aware joint training (\textit{Imb. Joint}), standard joint training (\textit{Std. Joint}), and 8 imbalance learning techniques on CIFAR-100 test set in the last session of a 5-way 5-shot FSCIL setting. The figure shows the base and incremental class accuracies of the top-5 trials for each method after 30 iterations of hyperparameter random search. The best trials are determined according to the \aacc. \textit{Imb. Joint} consists of CMO, Balanced Softmax, and ImbSAM.}
    \label{fig:imbalanced_top5}
\end{figure}

\myparagraph{Exploring Imbalanced Learning in FSCIL.} Figure~\ref{fig:imbalanced_top5} compares the base and incremental class accuracies across the top-5 trials for each method. The imbalance-aware joint training approach achieves markedly higher incremental accuracy, whereas most other approaches—including the standard joint training—show near-zero accuracy on incremental classes, indicating severe overfitting to base classes.

\begin{figure}[h]
    \centering
    \includegraphics[width=0.99\linewidth]{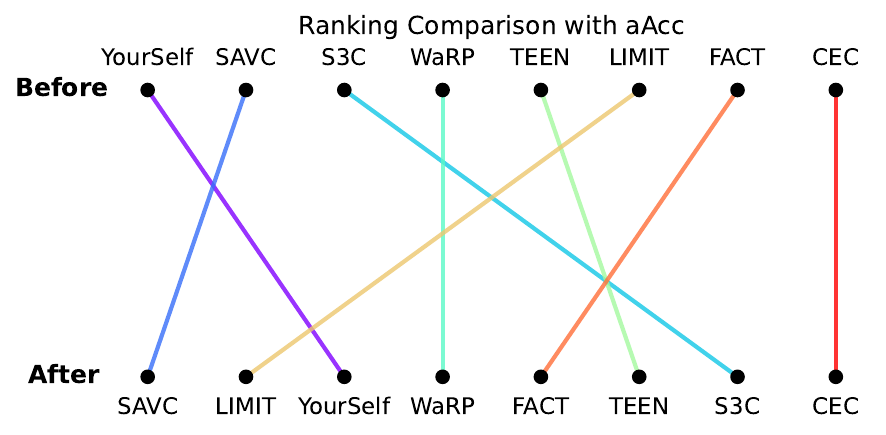}
    \caption{\small
    Comparison of FSCIL method performance rankings on CIFAR-100 test set before and after applying our standardized evaluation protocol. Rankings are presented in descending order from left to right.
    }
    \vspace{-1.em}
    \label{fig:kendall_corr_aacc}
\end{figure}

\begin{figure*}[t]
    \centering
    \subfloat[Base Classes]{
        \includegraphics[width=0.45\linewidth]{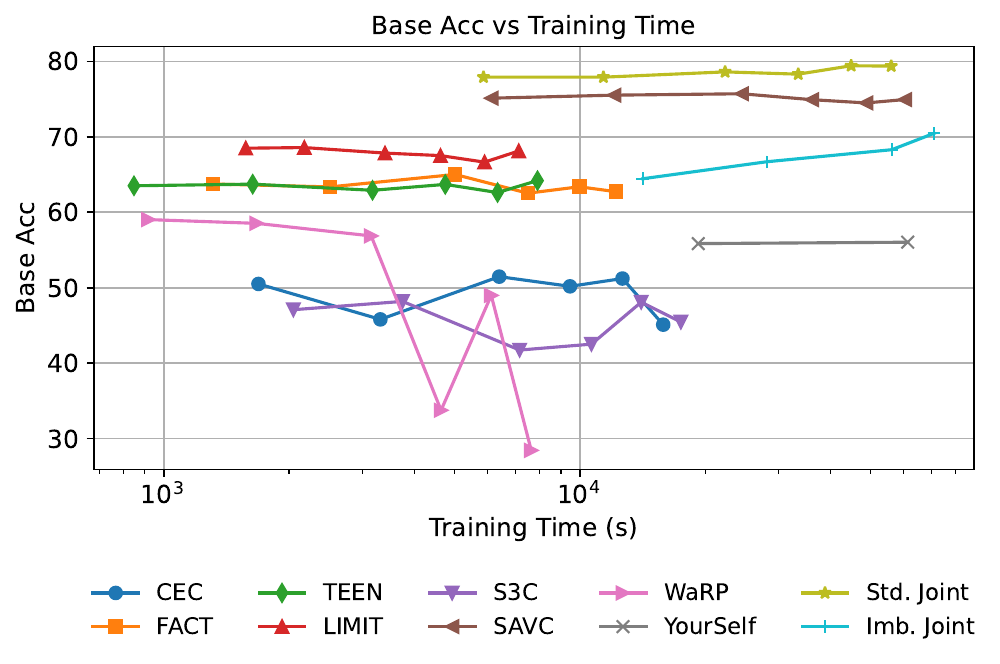}
        \label{fig:metric_base}
    }
    \hfill
    \subfloat[Incremental Classes]{
        \includegraphics[width=0.455\linewidth]{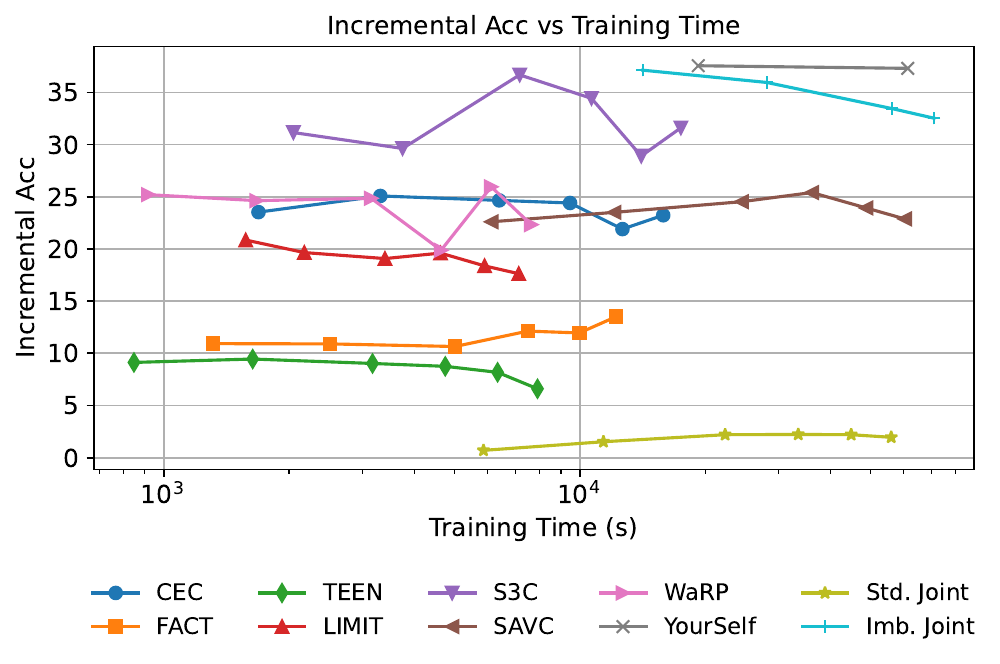}
        \label{fig:metric_incr}
    }
    \caption{\small
    Performance comparison of base and incremental classes across different training times on CIFAR-100 test set. Training times are presented on a log scale. \textit{Std. Joint} and \textit{Imb. Joint} denote standard joint training and imbalance-aware joint training, respectively.
    }
    \label{fig:base_incr}
\end{figure*}

\myparagraph{Performance Rankings Before and After
Protocol Standardization.} Figure~\ref{fig:kendall_corr_aacc} illustrates the ranking shifts of existing FSCIL methods in terms of \aacc before and after applying our standardized evaluation protocol. The “\textbf{Before}” rankings are based on the values reported in prior studies, whereas the “\textbf{After}” rankings reflect the results reproduced under our proposed protocol. All methods except for CEC and WaRP show changes in ranking. Particularly, S3C and LIMIT exhibit the largest shifts, each moving by four positions. These ranking shifts highlight inconsistencies in previous FSCIL experiments and evaluation, underscoring the need for a standardized protocol in future studies.

\vspace{1.em}

\myparagraph{Base and Incremental Performance by Training Time.} Figure~\ref{fig:base_incr} shows the \aacc of base and incremental classes with respect to training time across different methods. For base classes, standard joint training and SAVC achieve the highest performance, benefiting from their longer training durations. In contrast, YourSelf and imbalance-aware joint training mark the best results for incremental classes. Overall, imbalance-aware joint training maintains strong performance on both the base and incremental classes. Notably, TEEN and LIMIT exhibit a trend in which extended training improves base class performance but degrades incremental class performance, likely due to overfitting to base classes as the number of training epochs increases.

\end{document}